# KULTURE Bench: A Benchmark for Assessing Language Model in Korean Cultural Context

Xiaonan Wang[1], Jinyoung Yeo[2], Joon-Ho Lim[3], Hansaem Kim[1*]

[1]Interdisciplinary Graduate Program of Linguistics and Informatics, Yonsei University
[2]Department of Artificial Intelligence, Yonsei University
[3]Tutorus Labs, Republic of Korea

{nan, jinyeo, khss}@yonsei.ac.kr, jhlim@tutoruslabs.com

## Abstract

Large language models (LLMs) have exhibited significant enhancements in performance across various tasks. However, the complexity of their evaluation increases as these models generate more fluent and coherent content. Current multilingual benchmarks often use translated English versions, which may incorporate Western cultural biases that do not accurately assess other languages and cultures. To address this research gap, we introduce KULTURE Bench, an evaluation framework specifically designed for Korean culture that features datasets of cultural news, idioms, and poetry. It is designed to assess language models' cultural comprehension and reasoning capabilities at the word, sentence, and paragraph levels. Using the KULTURE Bench, we assessed the capabilities of models trained with different language corpora and analyzed the results comprehensively. The results show that there is still significant room for improvement in the models' understanding of texts related to the deeper aspects of Korean culture.

## 1 Introduction

The proliferation of LLMs capable of generating fluent and plausible responses has significantly complicated the task of assessing their knowledge and reasoning abilities (Li et al., 2024). To address this challenge, researchers have developed a range of benchmarks designed to assess diverse capabilities of models such as GLUE (Wang et al., 2018) and MMLU (Hendrycks et al., 2020). Although these benchmarks are primarily in English, when addressing multilingual issues, current evaluation practices often depend on translations of those datasets originally in English (Shi et al., 2022). However, a critical issue arises as their inherent Western cultural bias makes them unsuitable and inappropriate for evaluating LLMs across a variety of cultures and languages. Therefore, there has been a notable trend toward the development of culturally specific benchmarks that cater to local contexts such as IndiBias (Sahoo et al., 2024) for Indian culture, Heron-Bench (Inoue et al., 2024) for Japanese culture, and CCPM (Li et al., 2021) and ChID (Zheng et al, 2019) for Chinese culture. However, comparable research in Korea is still relatively scarce.

To bridge this resource gap, we developed KULTURE Bench, which includes 3584 instances across three datasets. This benchmark not only evaluates cultural knowledge but also includes datasets that inherently feature idioms and poetry, enriching the linguistic elements with deep cultural significance. Additionally, it thoroughly assesses language models' cultural comprehension and reasoning abilities across word, sentence, and paragraph levels.

We conducted evaluations on models trained with various primary corpora, including Clova X (Korean), GPT-4 and GPT-3.5 Turbo (English), and Tiangong (Chinese). The evaluation results show that, overall, the native Korean model, Clova X, performed the best, followed by GPT-4. However, even the highest-performing Clova X achieves only a 69.12% accuracy on the dataset containing idiomatic expressions, and the best result for the dataset featuring poetry is a mere 45.7% accuracy by GPT-4, which suggests that LLMs still face challenges when processing texts embedded with deep cultural and historical contexts. Further analysis of LLMs' performance on the KULTURE Bench reveals deficiencies in handling texts related to Korean culture and provides important insights for improvements in this area. KULTURE Bench can be accessed at `https://github.com/robot507/KULTUREBench.git`

---

*Corresponding Author

## 2 Related Work

### 2.1 Language Model

With the advent of Transformers and their self-attention mechanisms, the development of large English language models has accelerated rapidly. Models such as GPT-4 (OpenAI, 2023) represent notable achievements in natural language processing. Meanwhile, language models in other languages are also striving to bridge the gap with English. Examples include China's GLM (Zeng et al., 2022), Korea's HyperCLOVA (Kim et al., 2021), and Japan's Japanese StableLM (StabilityAI, 2023), which are all making significant advancements in their respective linguistic domains. With the advancement of multilingual language models, determining the appropriate methods for evaluating their language-specific capabilities becomes essential. This emphasizes the need for benchmarks tailored specifically to assess the multilingual proficiency of LLMs.

### 2.2 Korean Cultural Benchmarks

Several Korean culture-focused datasets have been developed as summarized in Table 1.

Jin et al. (2024) introduced the Korean Bias Benchmark for Question Answering, adapted from BBQ dataset (Parrish et al., 2022). This dataset builds on the original by adding new categories of bias specifically targeting Korean culture. HAE-RAE Bench (Son et al., 2023) is curated to evaluate large language models' understanding of Korean culture through six downstream tasks in vocabulary, history, general knowledge, and reading comprehension. CLIcK (Kim et al., 2024) sources its data from official Korean exams and textbooks, categorizing the questions into eleven subcategories under the two main categories of language and culture. This dataset challenges models' understanding of Korean culture through QA pairs. KorNAT (Lee et al., 2024) is launched to assess models' adherence to the distinctive national attributes of South Korea with a focus on social norms and shared knowledge.

These existing datasets typically use a question-and-answer format to test models' ability to handle culturally relevant content, but do not directly incorporate elements that represent unique cultural phenomena into the dataset. Thus, we introduce KULTURE Bench that not only assesses cultural knowledge through news articles but also incorporates idioms and poetry, which, originating from ancient times, remain widely prevalent in contemporary Korean society. This benchmark challenges models more rigorously by demanding a deeper understanding and contextual integration of enduring cultural elements.

| Benchmarks | Instances | Type |
|---|---|---|
| KoBBQ | 76048 | Bias |
| HAE-RAE BENCH | 1538 | Cultural Knowledge |
| CLIcK | 1995 | Cultural Knowledge |
| KorNAT | 10000 | Cultural Alignment |
| KULTURE Bench (Ours) | 3584 | Cultural Comprehension |

Table 1 Overview of Korean cultural benchmarks

## 3 KULTURE Bench

### 3.1 Task Overview

KULTURE Bench is an integrated collection comprising three datasets with a total of 3,584 instances, crafted to evaluate models' comprehension of Korean culture at different linguistic levels. The first dataset KorID features 1,631 instances and targets the understanding of Korean idioms through a cloze test format, where idioms are replaced with blanks that models must fill by interpreting the extended meaning from selected candidates, assessing word-level cultural insights. The second dataset KorPD contains 453 instances. In this dataset, a specific line from a Korean poem is replaced by a blank, requiring the model to infer the correct line based on the poem's overall meaning, rhythm, and rhetorical style from a set of selected candidate lines. This dataset is designed to evaluate the sentence-level cultural comprehension of the models. The third dataset KorCND contains 1,500 instances. Here, models are tasked with summarizing culturally relevant Korean news articles and selecting the correct headline from a set of designed candidates. This dataset focuses on the models' ability to comprehend and summarize extended texts, reflecting their paragraph-level understanding of Korean culture. Appendix A provides examples from three datasets in the KULTURE Bench.

In selecting the source materials for these datasets, we prioritized real-world content like current news, common idioms, and classical poems from Korean textbooks for dataset relevance and authenticity.

## 3.2 KorID: A Korean Four-character Idiom dataset for Cloze Test

### 3.2.1 The characteristics of four-character idioms

Four-character idioms are widely used in Asian countries such as China, Japan, Vietnam, and Korea. Four-character idioms, known as Sajaseong-eo or Hanjaseong-eo in Korea, are a significant part of the Korean language and originated from ancient China; over history, they were transmitted to the Korean Peninsula where new idioms incorporating Korean cultural elements also emerged locally. To this day, idioms remain a frequently used linguistic phenomenon in Korean society.

Many idioms feature metaphorical meanings that stem from the stories behind them, meaning that their true meanings cannot be deduced simply by interpreting the literal meanings of the words they are composed of. For example, the idiom "함흥차사" literally means "an envoy sent to Hamhung" (Hamhung: a location on the Korean Peninsula), but its metaphorical meaning refers to situations where someone sent on an errand has not returned any news, or responses are significantly delayed. This idiom originates from an early Joseon Dynasty story about an envoy who was sent to Hamhung to escort the founding monarch, Lee Seong-gye, but never returned. Therefore, understanding idioms requires knowledge of their underlying stories and culture and using cloze tests with idioms can assess a model's ability to comprehend and apply the cultural nuances and metaphorical meanings embedded within the language.

### 3.2.2 Construction of KorID dataset

KorID dataset is constructed in four main stages: 1. Vocabulary Construction, 2. Passage Extraction, 3. Candidates Retrieval, 4. Synonym Check.

**Idiom Vocabulary Construction** The idiom vocabulary for the KorID dataset was derived from two primary sources: *Gosa, Saja Four-character Idiom Grand Dictionary* (Jang, 2007), known for its extensive collection of over 4,000 idioms, and *Korean Four-Character Idiom Grand Dictionary* (Han, 2011), which includes around 2,300 idioms originating from the Korean Peninsula and approximately 270 Chinese idioms used in Korean classics.

After performing OCR on the idiom lists within the dictionaries, the four-character idioms were digitized and stored in Excel. Each entry underwent a manual review and correction process to fix any OCR recognition errors, with strict adherence to the four-character criterion. An automated script was then employed to remove duplicates from the collected idioms, ultimately yielding a total of 5,372 unique idioms.

**Passage Extraction** This research utilized the Modu Corpus[1], which includes several versions of the NIKL Newspaper Corpus, containing a comprehensive collection of Korean news articles from 2009 to 2022, sourced from a wide range of national and regional outlets and organized in JSON format.

Using a Python script, the process of matching idioms with relevant news passages was automated, systematically searching the news corpus for occurrences of each of the 5,372 four-character idioms from the vocabulary construction phase. Whenever an idiom appeared in a news article, the article was stored in an Excel file next to the idiom it contained.

The idioms selected through this method are frequently used in daily news, while other less common idioms were removed. This automated search and storage resulted in an Excel file where each row contains an idiom followed by the news text that includes it.

After searching the entire news corpus, 1,631 instances where idioms from the vocabulary appeared in the news were identified. During text preprocessing, special attention was given to refining the news text extracted, involving the removal of unnecessary elements like reporter's email addresses and extraneous metadata, which do not contribute to linguistic analysis or the cloze test structure.

Additionally, any Hanja (Chinese character) explanations before or after the idioms within the text were also removed to maintain focus on the relevant textual content.

---
[1] https://kli.korean.go.kr/corpus/main/requestMain.do

**Candidates Retrieval** This part involves measuring the semantic relevance between idioms using cosine similarity of their embeddings obtained from the KorBERT model (Lim et al. 2020) developed by ETRI. Each target idiom in a news text is compared with others in the idiom vocabulary list to calculate cosine similarity, defined as:

$$\text{cosine similarity} = \frac{\mathbf{A} \cdot \mathbf{B}}{\|\mathbf{A}\| \cdot \|\mathbf{B}\|}$$

From the results, one idiom is chosen from each of four predefined cosine similarity ranges: [0.5, 0.6], [0.61, 0.7], [0.71, 0.8], and [0.81, 0.9], resulting in four candidates. Together with the target idiom, labeled as the "Golden Answer," five options are generated for each cloze test. These idioms replace the target idiom's position, now a blank, in the news text.

**Synonym Check** Four-character idioms are a category of words that possess synonyms, allowing for interchangeable use in some contexts. To ensure the integrity of the KorID dataset, a review process was conducted. Fifty samples from the dataset were randomly selected and manually inspected to determine whether, aside from the Golden Answer, other terms could also meet the standards for insertion into the texts.

The review standards were as follows: (1) two idioms are nearly identical in both literal and deeper meanings and can be interchangeably used in the given news text (3 points); (2) the idioms share similar themes or elements but have distinct differences in meaning (2 points); the idioms are fundamentally different in meaning (1 point).

Following this review, it was found that all the Golden Answers across the 50 samples possessed a unique selectivity. This outcome validates our strategy of excluding potential candidates with a similarity score above 0.9, demonstrating the effectiveness of using similarity scores to select negative options during the dataset creation process. Following these steps, the KorID dataset is constructed.

### 3.3 KorPD: A Korean Poem Dataset for Cloze Test

#### 3.3.1 The characteristics of poems

Distinct from other literary forms, poetry is known for its intense emotions, clear stylistic expression, and abstractly conveyed rich themes. The semantics in poetry are often more ambiguous and complexly intertwined than in other literary forms (Li et al., 2021), which complicates the automatic analysis and evaluation of poetic semantics, thus necessitating more efforts in assessing language models' understanding of poetic meaning.

#### 3.3.2 Construction of the KorPD dataset

KorPD dataset is constructed in two main stages: 1. Poem Collection, 2. Candidates Retrieval.

**Poem Collection** Poems in KorPD dataset were compiled from Korean textbooks. To collect these poems, OCR was employed to digitize content directly from textbooks, which was then systematically organized into an Excel file with columns for poem titles, poets' names, and the content of each poem to facilitate access and further processing.

Following digitization, a thorough manual review ensured the accuracy and integrity of the data, with corrections made to rectify any errors detected. From these efforts, a total of 91 poems were selected, offering a rich repository of poetic content that reflects contemporary educational and cultural contexts in Korea, encompassing a broad range of themes and styles.

**Candidates Retrieval** Using a set of 91 Korean poems, a specific line is randomly extracted from each poem using a Python script, repeated up to five times for poems with more than five lines, generating 453 unique items. Each line is only selected once across the dataset.

For semantic analysis, the [CLS] token's hidden state from ETRI's KorBERT model (Lim et al. 2020) serves as the semantic representation of each line, capturing its contextual essence crucial for accurate comparisons.

Cosine similarity then quantifies the semantic resemblance between the golden answer and other lines, grouping them into four similarity intervals: [0.5-0.6], [0.61-0.7], [0.71-0.8], and [0.81-0.9]. From each interval, one line is randomly chosen to join the golden answer in the dataset item, ensuring the distractors are similar enough to challenge identification yet diverse enough to span a broad semantic spectrum.

The order of the golden and negative answers is randomized to increase the dataset's testing robustness. Following these steps, the KorID dataset is constructed.

## 3.4 KorCND: A Korean Culture News Dataset for Headline Matching

### 3.4.1 The characteristics of news headline

News headlines skillfully employ dynamic verbs and vivid adjectives to quickly transmit essential messages, often integrating rhetorical devices like ellipsis and inversion to maximize space efficiency and reader engagement. For example, the Korean headline "서울시, 문화와 함께하는 추석… 신달자·정호승의 시낭독, 서울시향 연주" (Seoul, Celebrating Chuseok with Culture... Poetry Readings by Shin Dal-ja and Jeong Ho-seung, Performance by the Seoul City Symphony) not only introduces a cultural event but also encapsulates the broader cultural context of Chuseok. To fully comprehend such headlines, language models need to interpret beyond the words to grasp cultural and contextual nuances. This involves understanding the celebration of Chuseok, a major Korean festival, which includes traditional customs like family reunions and specific foods such as "송편" (songpyeon, rice cakes). The model must recognize the headline's reference to specific cultural activities and significant figures in Korean arts, linking these elements to convey the depth of the festival's cultural significance accurately. Such capabilities are essential for models to interpret the text effectively and align their responses with culturally relevant narratives.

### 3.4.2 Construction of the KorCND dataset

KorCND dataset is constructed in two main stages: (1) News Collection, (2) Candidates Retrieval.

**News Collection** The KorCND Dataset primarily sources its texts from the NIKL Newspaper Corpus 2022 (v1.0) within the Modu Corpus [2]. This extensive resource includes 978,342 newspaper articles from 2021, gathered from 34 different news outlets and covering a wide range of topics such as culture, society, economy, and sports, all provided in JSON format.

The dataset was created by initially filtering these articles for cultural relevance, using keyword and tag-based searches to identify those related to "문화" (Culture). Following this automated selection, a manual review was conducted to ensure the articles were specifically pertinent to Korean culture, leading to the curation of 1,500 articles that effectively reflect the diversity of Korean cultural life.

Once the article selection was finalized, the next step was to organize the chosen articles into a structured format suitable for in-depth analysis by compiling the news titles and corresponding text content into an Excel file. This format was chosen for its accessibility and ease in data manipulation and review. In the spreadsheet, Column A was designated for news headlines, providing a clear reference, and Column B stored the full text of the articles, ensuring all textual information was preserved and easily accessible.

Quality control measures were implemented to maintain the integrity and quality of the data during this organization phase. Each entry was meticulously checked for data entry errors, such as misaligned text or incomplete headlines.

**Candidates Retrieval** After collecting and categorizing news headlines and their texts into the KorCND Dataset, the next step involved meticulously extracting negative choices to enhance the dataset's utility in assessing language models, particularly for tasks that require distinguishing between closely related texts. The aim was to select opposing candidates that closely resemble the Golden Answer to test the model's precision in identifying subtle differences.

The embedding of news headlines was performed using the KorBERT model developed by ETRI, tailored specifically for the Korean language to capture deep semantic meanings. The semantic representation of each headline was extracted from the hidden state of the [CLS] token of the KorBERT model (Lim et al. 2020), which is designed to encapsulate the overall meaning of the input sequence.

Once embeddings were derived from the [CLS] tokens, the similarity between different headlines was quantified using the cosine similarity metric, focusing on nuanced semantic relationships over mere lexical similarities for a more precise similarity assessment.

In determining effective negative choices, each Golden Answer's [CLS] token representation was compared against all others in the dataset. This comparison yielded similarity scores for each headline pair, facilitating the identification of headlines that are similar enough to potentially confuse the model but distinct enough to test its discriminative power. Headlines were then sorted into predefined similarity ranges: [0.5, 0.6], [0.61, 0.7], [0.71, 0.8], and [0.81, 0.9], each representing varying levels of difficulty in terms of semantic closeness.

---

[2] https://kli.korean.go.kr/corpus/main/requestMain.do

From each similarity interval, one headline was randomly selected as a negative choice. This selection not only introduced necessary variability but also simulated realistic scenarios where language models must discern between semantically similar phrases. This methodical approach ensured the dataset not only challenged but also accurately evaluated the nuanced understanding and processing capabilities of contemporary language models.

Finally, the order of the Golden Answer and the Negative Answers was randomized. Following these steps, the KorCND Dataset is constructed.

## 4 Experimental Setup

### 4.1 Evaluated Models

We assessed models trained primarily on distinct language corpora including GPT-4[3] and GPT-3.5 Turbo[4] (English), Clova X[5] (Korean).

We also tested Tiangong[6], a model focuses on Chinese, aims to explore potential spillover effects, given the historical influence of Chinese Hanja on the Korean language.

### 4.2 Prompt Types

We designed two rounds of experiments to assess the ability of language models to process complex texts. In the first round, we employed the Zero-shot Prompting method to directly evaluate the comprehension of complex texts by four selected models. In the second round, to investigate whether the Chain of Though technique could enhance model accuracy in cultural contexts and to analyze the impact of reasoning length on model accuracy, we required models to answer after a period of contemplation and set three different reasoning length requirements: 1-2 sentences (short reasoning), 3-4 sentences (medium reasoning), and 5-6 sentences (long reasoning). The second round of experiments was specifically conducted using the KorID dataset on three models: GPT-3.5 Turbo, Clova X, and Tiangong. Specific prompt settings are provided in Appendix B .

### 4.3 Evaluation Metric

We employ accuracy as the evaluation metric. This means calculating the proportion of samples for which the model provides the correct answer out of the total number of samples. The higher the accuracy, the better the performance of the model.

### 4.4 Response Validation Criteria

During the experiment, it was observed that models occasionally produce verbose responses. To accurately extract the selected answers from the models' responses, the following acceptance criteria were established: The answer must i) exactly match a term provided in the options, ii) include specific expressions clearly intended to convey the answer, such as "the answer is -", iii) present the option enclosed in square brackets [].

Responses that do not meet these criteria are considered out-of-option answers.

## 5 Results

### 5.1 Main Results

**Model Comparison** According to Table 2, Clova X, which focuses on Korean, is observed to be the top-performing model, achieving an accuracy of 69.12% on the KorID dataset and an impressive 92.77% on the KorCND dataset, ranking first in both. Following closely, GPT-4 also shows strong performance, even surpassing Clova X with an accuracy of 45.7% on the KorPD dataset. In contrast, GPT-3.5 Turbo underperforms compared to GPT-4, further demonstrating the superior capabilities of GPT-4 as a more advanced iteration. Tiangong records the lowest accuracy, indicating that mere cultural similarities are insufficient to ensure outstanding model performance.

**Dataset Comparison** According to Figure 1, we observe that all models exhibit relatively high performance when processing news texts, but significantly lower accuracy with texts containing rich cultural elements such as poetry and idioms. We believe this is due to two main reasons: firstly, there is a scarcity of such data in the training corpora; secondly, these texts are laden with cultural information, which increases the complexity of processing.

---

[3] https://openai.com/index/gpt-4/
[4] https://platform.openai.com/docs/models/gpt-3-5-turbo
[5] https://clova-x.naver.com/welcome
[6] https://model-platform.tiangong.cn/overview

| Models | KorID | KorPD | KorCND |
|---|---|---|---|
| Tiangong | 26.92 | 14.57 | 70.61 |
| GPT3.5-Turbo | 31.58 | 25.61 | 77.65 |
| GPT-4 | 51.50 | **45.70** | 89.68 |
| Clova X | **69.12** | 37.75 | **92.77** |

Table 2 Accuracy (%) of different models on KorID, KorPD and KorCND datasets

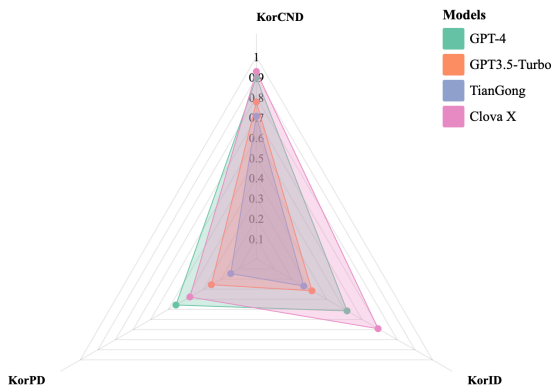

Figure 1 Models' performances on different datasets in KULTURE Bench

## 5.2 Analysis

**Does Chain of Thought Enhance Understanding and Reasoning in Culture Texts?**

As shown in Table 3, the Chain of Thought (CoT) strategy improved the accuracy of the Tiangong and Clova X models. However, the performance of GPT3.5-Turbo did not show notable improvement and sometimes even declined. Clova X, which is primarily trained on Korean language corpora, has a deep understanding of Korean culture; the Tiangong model, based on Chinese corpora, benefits from the cultural proximity between China and Korea; whereas GPT-3.5 Turbo, trained on English corpora, is more distanced from Korean culture. We speculate that a lack of appropriate cultural background knowledge may lead to inaccurate or erroneous reasoning when models execute tasks. To further validate this hypothesis, we randomly selected 50 correct responses from each model's response pool in the CoT experiment for an in-depth analysis, totaling a review of 150 responses. The analysis primarily examined whether the models correctly used relevant cultural knowledge in the reasoning process to select the correct answers, or if the correct answers were merely the result of random selection by the models. The analysis showed that the reasoning process of the Clova X model was correct 96% of the time. The Tiangong model reasoned correctly in 74% of the cases, while GPT-3.5 Turbo only demonstrated correct reasoning in 46% of instances. This indicates that training with Korean language corpora and possessing extensive knowledge of Korean culture makes the reasoning process of the Clova model effective. Analysis of Tiangong's responses revealed that 90% of the idioms correctly reasoned by this model are still in active use in contemporary Chinese society (based on the observation of whether the target idiom exists in online Chinese idiom dictionary [7]), thereby supporting the correct answer choices. These findings support the assumption that without sufficient cultural knowledge, the reasoning capabilities of models do not significantly improve and may even be impaired by incorrect cultural interpretations

Additionally, based on Table 3 and Figure 2, it can be observed that the accuracy of the TianGong model starts at 26.92%, rises to 35.83% with short reasoning, but then decreases to 33.70% and 32.88% with medium and long reasoning, respectively. This suggests that excessive reasoning may lead to decreased accuracy due to potential information overload. Clova X's accuracy begins at 69.12%, increases to 71.28% with short reasoning, but as the reasoning extends, it drops to 70.69% and 67.80%. This implies that properly calibrated CoT steps can enhance accuracy, but too many may negatively impact performance.

| Model | No CoT | Short Reasoning | Medium Reasoning | Long Reasoning |
|---|---|---|---|---|
| **Clova X** | 69.12 | 71.28 | 70.69 | 67.80 |
| **GPT3.5-Turbo** | 31.58 | 30.72 | 34.83 | 31.53 |
| **Tiangong** | 26.92 | 35.83 | 33.70 | 32.88 |

Table 3 Accuracy (%) of different models on the different ways of using Chain of Thought

---
[7] http://www.guoxue.com/chengyu/CYML.htm

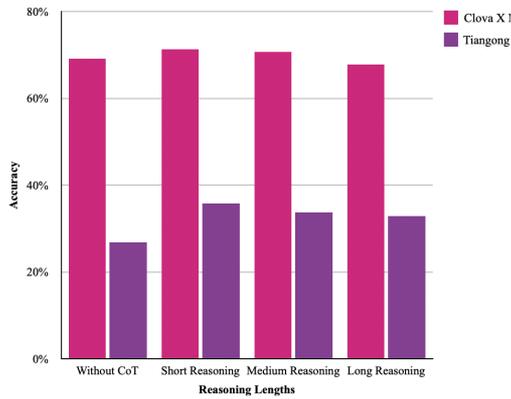

Figure 2 Accuracy (%) given by Tiangong and Clova X when the length of reasoning process changes

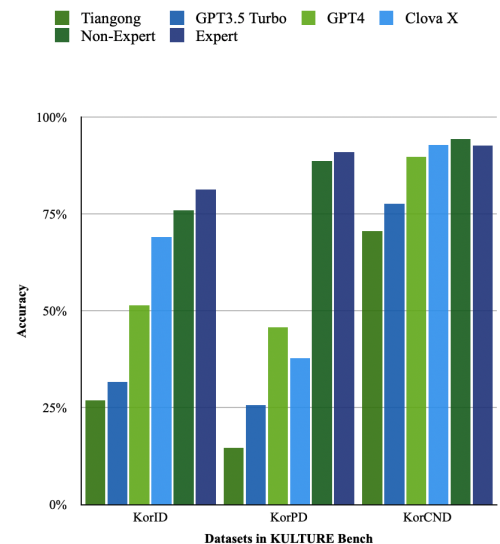

Figure 3 Model and human performance comparison across KULTURE Bench datasets

**How Do Model Performances Compare to Human Level Proficiency?**

Participants were split into two groups based on their academic backgrounds: 20 graduate and doctoral students in Korean Studies, termed "experts," and 20 undergraduate students without a Korean Studies focus, called "non-experts." Each participant received a coffee voucher as compensation for their time. For evaluation, 15 questions were randomly selected from each of the KULTURE Bench datasets.

As shown in Figure 3, there are significant differences between LLMs and human participants, with humans outperforming LLMs in both expert and non-expert groups, especially in handling cultural news and poetry. While LLMs perform well in structured tasks like news headline matching, they struggle with the complexities of cultural and linguistic contexts in poetry and idiomatic expressions, where humans, particularly those with specialized knowledge, excel.

**What Types of Errors Occur in Model Responses to Korean Cultural Texts?**

To investigate the types of errors in automated responses to Korean cultural texts, each model was required to explain its reasoning in a second-round conversation after initial evaluations. For analysis, 20 error responses were extracted from each model across three datasets, resulting in 60 samples per model, totaling 240 error samples across all models.

The analysis categorizes the errors in model responses into five types: (1) semantic understanding errors, (2) lack of cultural background knowledge, (3) grammatical or lexical errors, (4) logical errors, and (5) insufficient context interpretation. Appendix C provides specific examples of these types of errors.

## 6 Conclusion and Limitations

We introduce KULTURE Bench, a comprehensive dataset ranging from words to paragraphs that focuses on Korean cultural knowledge and linguistic phenomena, sourced from Korean news, textbooks, and dictionaries. Through our analysis and experiments, we observed that models perform best on modern texts and news but lack proficiency in idioms and poetry. We also found that without sufficient cultural knowledge, CoT techniques cannot be effectively utilized. Additionally, even when CoT is effective, inappropriate reasoning lengths can impact model accuracy. Moreover, human capabilities in KULTURE Bench task still surpass those of the models. The limitations of this research include the need to evaluate a broader range of models for a more comprehensive assessment. Additionally, the CoT experiments focused solely on cultural reasoning in Korean, and it would be beneficial to test whether similar results occur in other languages and cultures. Furthermore, the human capability assessment was limited by a small sample size, and there is a need to expand the sample for more comprehensive testing.

## A  Examples from KULTURE Bench

Table 4 shows an example from KorID dataset. In this dataset, models are required to read news articles, identify the blanks where idioms are missing, understand the meaning of each idiom presented in the options, and select the appropriate idiom based on the context.

Table 5 shows an example from KorPD dataset. In this dataset, models are required to read classical poems, identify the missing line, interpret the meaning of each line presented in the options, and then, considering factors such as the poem's meaning, emotional expression and rhythm, select the correct answer.

Table 6 shows an example from KorCND dataset. In this dataset, models are tasked with discerning the main idea of culturally relevant news articles, understanding the cultural elements present in the news, and selecting a headline from the options that accurately summarizes the content of the news.

## B  Prompt Types

Figure 4 shows the prompts used in the first round of experiments to evaluate language models; Figure 5 shows the prompts used in the second round of Chain of thought experiments.

**Experiment on KorCND**
Given the text: '{content}'. Here are the options: {', '.join(options)}.
Read the content carefully and choose the most appropriate headline from the options and mark the answer you chose with [].

**Experiment on KorID**
Given the text: '{content}'. Here are the options: {', '.join(options)}.
Select only one four-character idiom that fits best for the placeholder #idiom# in the content and mark the answer you chose with [].

**Experiment on KorPD**
Given the poem: '{content}'. Here are the options: {', '.join(options)}.
There is a missing line represented by #sentence# in the poem, choose the best line from the provided options to complete the poem and mark the answer you chose with [].

Figure 4 Prompts used in the first-round experiment

| | |
|---|---|
| Passage & Blank | …박곡리 노인회는 평균 연령 83세 이상으로 최고령 박순득 어르신여96을 비롯해 100여 명의 회원으로 구성돼 있으며 7가구는 독거노인이다. 집에 있는 흔한 닭은 하찮게 생각하고 멀리 있는 꿩은 귀하게 생각한다는 #idiom#란 말을 곱씹어본다. 최희동 박곡리 이장은 곁에서 부모님을 모시면서 이웃어른을 보살피는 박곡리 부녀회와 청년회가 오히려 멀리 떨어져 있는 친자식들보다 더 효자효부 노릇을 한다는 평소 어르신들의 말씀이 결코 낯설지 않다고 칭찬했다…（English Translation: "The senior citizens' association in Bakgok-ri has an average age of over 83, including the eldest member, Soon-deuk Park, aged 96, and comprises about 100 members, of whom seven are elderly living alone. One ponders the #idiom#, 'The common chicken at home is taken for granted, while the pheasant in the distant field is prized.' Hee-dong Choi, the village head of Bakgok-ri, praised the local women's and youth associations for caring for elderly neighbors and their own parents, saying that they often fulfill their filial duties better than children who live far away, a sentiment that is not unfamiliar to the elders." |
| Option Golden Answer | 가계야치 (Common things are regarded as lowly, while distant things are considered precious.) |
| Options (Negative) | 빈부귀천 (wealth and poverty, nobility and lowliness) 설상가상 (to make a bad situation even worse.) 무위이화 (Seemingly inactive, yet transforming.) 어동육서 (Ritual customs: Fish east, meat west) |

Table 4 An example from KorID dataset

| | |
|---|---|
| Poem & Blank | 나 두 야 간다<br>나의 이 젊은 나이를<br>눈물로야 보낼 거냐<br>나 두 야 가련다<br>#sentence#<br>안개같이 물 어린 눈에도 비치나니<br>골짜기마다 밭에 익은 묏부리모양<br>주름살도 눈에 익은 아- 사랑하던 사람들<br>버리고 가는 이도 못 잊는 마음<br>쫓가는 마음들 무어 다를 거냐<br>돌아다보는 구름에는 바람이 희살짓는다<br>앞 대일 언덕인들 마련이나 있을 거냐<br>나 두 야 가련다<br>나의 이 젊은 나이를<br>눈물로야 보낼 거냐<br>나 두 야 간다.<br>I too shall leave<br>My youthful days<br>Must they be spent in tears?<br>I am indeed pitiable<br>#sentence#<br>Even in eyes wet as mist, reflections can be seen<br>In every valley, the familiar shape of grave mounds underfoot<br>Familiar wrinkles, ah- those beloved people<br>Hearts that cannot forget even those who leave Chasing hearts, how different are they?<br>The wind plays in the clouds looking back<br>Are there hills ahead just as supposed?<br>I am indeed pitiable<br>My youthful days<br>Must they be spent in tears?<br>I too shall leave. |
| Option Correct Answer | 아늑한 이 항구-ㄴ들 손쉽게야 버릴 거냐 (Can these cozy harbors be so easily abandoned?) |
| Options | 그날이 와서, 오오 그날이 와서 (That day comes, oh, that day comes.) 고요히 다물은 고양이의 입술에 (On the quietly closed lips of a cat.) 밤이면 실컷 별을 안고 (At night, embracing the stars to the fullest.) 백마 타고 오는 초인이 있어 (There is a superman coming on a white horse.) |

Table 5 An example for KorPD dataset

## Chain-of-Thought Experiment on KorID

Given the text: '{content}'. Here are the options: {', '.join(options)}. Please analyze in one to two sentences and then select the appropriate idiom from options to fill in the blank which is marked as #idiom# in the text, and mark the answer you chose with [].

Given the text: '{content}'. Here are the options: {', '.join(options)}. Please analyze in three to four sentences and then select the appropriate idiom from options to fill in the blank which is marked as #idiom# in the text and mark the answer you chose with [].

Given the text: '{content}'. Here are the options: {', '.join(options)}. Please analyze in five to six sentences and then select the appropriate idiom from options to fill in the blank which is marked as #idiom# in the text and mark the answer you chose with [].

Figure 5 Prompts used in the second-round (Chain of Thought) experiment

| | |
|---|---|
| News Article | 맹숙영 시인이 시집 '우리가 사랑할 수 있는 시간' (황금마루)를 펴냈다. 맹 시인의 7번째 시집이고 신작 70여편을 수록했다. 이인평 산림문학 편집주간은 이 책의 추천사에서 "맹숙영의 시편에는 살아온 세월을 반추하면서 언어로 다듬어낸 그만의 목소리가 담겨있다. 세상을 향해 애틋하게 바라본 시각이 그의 숨결로 다가오기 때문" 이라고 했다. 문학평론가 이덕주 시인은 평설을 통해 "이번 시집의 특성은 시인의 사색적 성찰이 균형있게 내장돼 시인의 진솔한 자기 증언을 확인할 수 있다는 점" 이라며 "이런 관점은 우리 앞에서 전개되는 세상에 대해 독창적인 해석을 내리기 때문에 용인된다. 개별적 차별보다 존재마다 다양성을 존중하는 화합의 정신, 즉 지극한 사랑을 전제했기에 가능했을 것" 이라고 했다. 맹 시인은 성균관대 영어영문학과를 졸업하고 영어 교사를 지냈다. 한세대 사회복지대학원도 졸업했다. 한국창조문학으로 등단한 그는 공간시낭독회 상임시인, 한국문인협회 한국시문학아카데미 푸른초장문학회 신문예학회 사월회 회원이다. 바람칼의 칸타빌레 동인이기도 하다. 창조문학대상, 양천문학상, 한국기독시문학상 등을 수상했다. 양천문학 부회장을 역임하고 자문위원으로 활동 중이다. 좋은시공연문학 한국크리스천문학회 부회장, 한국창조과학문학 운영 이사, 한국현대시인협회와 기독시인협회 이사 등을 맡고 있다. 시집으로 '사랑이 흐르는 빛', '꿈꾸는 날개', '바람 속의 하얀 그리움' 한영 대역 '불꽃 축제', '아직 끝까지 않은 축제', '아름다운 비밀' 등이 있다. (English Summary: Maeng Suk-young, a distinguished poet, has released his seventh poetry collection titled 'The Time We Can Love', featuring around 70 new poems. The collection is praised for integrating reflective introspection that confirms Maeng's honest self-expression and offers an original perspective on the world, underpinned by a spirit of harmony and profound love. Maeng, a graduate of Sungkyunkwan University in English Language and Literature and Hansei University's Graduate School of Social Welfare, has a notable career as an English teacher and has earned several prestigious awards. He is actively involved in various literary and creative organizations, contributing significantly to the Korean literary scene.) |
| Option Golden Answer | ["천지에 꽃피고 꽃지고 나면 향기 가득…"] ("Flowers bloom and wither, filling the air with fragrance…") |
| Options Negative Answers | [윤범모 취임공약 한국 근현대 미술 120년사 나왔다] (Yoon Beom-mo's Inaugural Promise: 120 Years of Korean Modern and Contemporary Art Released) [신명난 국악이 함께하는 올해 마지막 한달...광주대표 브랜드 공연 '국악상설'](End-of-Year Vibes with Gwangju's 'Regular Gugak' Korean Music Shows) [사뿐사뿐~ 우아한 몸짓에 여름밤 황홀](Soft and Graceful Movements Enchant the Summer Night) ['자유민주시인상' 수상작 80편 담긴 '칼날 위에서 피는 꽃' 출간]('Freedom Democracy Poet Award' Winning 80 Pieces Featured in 'Flower Blooming on the Blade' Published) |

Table 6 An example from KorCND dataset

## C Categories of Errors in Model Responses

**Semantic understanding errors** occur when a model misinterprets meanings, nuances, or implications of words, phrases, or the overall context in a text. This error often involves misunderstanding the text's intent, emotional tone, or thematic connections, leading to responses that are logically coherent but contextually or factually incorrect. Table 7 provides an example of this error.

| | |
|---|---|
| News Text | 전 권투선수이자 시인인 홍영철씨가 새로운 시집 '이 땅에서 사랑하고, 로상까지!'(하나로선 사상과문학사)를 출간했다. 이 시집에는 그의 신작 94 편이 수록되어 있으며, 그의 삶은 마치 한 편의 드라마와 같다. 7남매 중 다섯째로 태어난 그는 어린 나이에 가족과 헤어지고 힘든 시절을 보냈다. 명문 서울고등학교에 진학했지만 경제적인 이유로 2 학년 때 중퇴했다. 이후 거리에서 주먹으로 생계를 유지하다가 권투를 시작했다. 홍씨는 한때 우울증과 공황장애를 겪었으나, 선수 생활 중 목회자를 만나 예수를 영접하고 새로운 삶을 시작했다. 그는 대통령배 복싱선수권 대회에서 우승하고 프로복싱 신인왕에 올랐다. 또한 서울 서대문의 태권도 관장과 정심관 이종투기 연내 관장을 역임했다. 현재는 프로스펙스 복싱 동우회 회장, 투혼 이종투기 협회 이사, 한국문인협회 회원, 힌우물 동인회 회원으로 활동 중이다. 그는 국제 킥복싱에서 최우수체육관상과 공로상을 수상했으며, 2019 년에는 '하나로선 사상과 문학상' 대상(시 부문)을 받았다. 그의 동인지로는 '막사발시', '한우물', '잔고개', '옹헤야', '시마루' 등이 있으며, 시집으로는 '아 달고나', '거리에 꽃피어라', '내 영혼이 우는 날까지', '한낱 스쳐가는 바람인가요', '달고나 장수' 등이 있다. (English Summary: Hong Young-cheol, a former boxer turned poet, has published a new poetry collection titled "Love in This Land, Even to the Grave." The collection includes 94 of his latest poems, reflecting his dramatic life journey from a troubled youth to a champion boxer and now a recognized poet. Hong's life, marked by overcoming adversity and finding faith, also includes notable achievements in both sports and literature, serving in leadership roles in martial arts and contributing to various literary communities.) |
| Model Choice (From GPT-4) | 국민 66% "문학책 1권도 안 샀다"… 작가 71% "문학 수입 월 50 만원 미만" (66% of the population "did not buy even one book of literature"... 71% of writers "earn less than 500,000 won per month from literary activities") |
| Golden Answer | 아름다운 소년의 마음 꼬옥 띄운다 (The beautiful boy's heart is gently sent off.) |
| Model Response | The headline '국민 66% "문학책 1 권도 안 샀다"… 작가 71% "문학 수입 월 50 만원 미만"' is the best choice. I noticed that the text involves a poet's personal experiences and literary achievements. The poet's diverse professional background and awards in the literary world led me to explore a broader literary market and social context. Therefore, I chose the title '국민 66% "문학책 1 권도 안 샀다"… 작가 71% "문학 수입 월 50 만원 미만"' This title reflects the widespread status of the literary market, especially concerning the public's purchasing behavior of literary works and the economic conditions of writers. I believe this title highlights the poet's perseverance and effort in facing economic challenges in his literary career, as mentioned in the text. In my view, this connection not only reveals the challenges of individual artists but also reflects the real predicaments of the entire literary world. |

Table 7 An example of semantic understanding error in model responses

This error involves a misinterpretation of the main theme and focus of the news text. The original article primarily discusses the personal life experiences, career transitions, and literary achievements of the poet. However, the title selected by the model focuses on the general economic conditions of the literary market, such as the purchase rates of literary works and the income levels of writers. The chosen title completely overlooks the humanistic and emotional aspects of the content, opting instead for a cold, statistical perspective. While the title provided by the model appears reasonable on the surface and relevant to the "literature" theme, it fails to accurately capture the essence and emotional tone of the original text. This type of semantic understanding error leads to an incorrect choice of news headline.

**Lack of cultural background knowledge** leads to errors where models make inappropriate choices due to insufficient understanding of cultural contexts. These errors can result in responses that are logically coherent but culturally inappropriate or inaccurate, reducing the authenticity and reliability of the model's output. Table 8 provides an example of this type of error.

The idiom "삼고초려" is used to describe the belief that one must be patient and wholeheartedly devoted to attracting outstanding talents. This term reflects the idea that high-status individuals should humbly and sincerely strive to bring those with exceptional abilities into their circle. However, when answering this question, the model, lacking the cultural knowledge behind "삼고초려," misinterpreted its meaning as "carefully considering and thinking deeply before making a decision," leading to an incorrect choice.

**A grammatical or lexical error** occurs when a language model makes mistakes in the syntax or the orthography of the language it is generating. These errors can significantly affect the clarity and professionalism of the text, leading to misunderstandings or reducing the credibility of the content. Table 9 provides an example lexical error. In this example, although the model's understanding of the idiom is correct and the choice is appropriate, it was judged to be incorrect because the correct Korean was not outputted in the response.

| News Text | 물 팔아 아시아 1등 됐다…마윈 제친. 블룸버그는 자사의 억만장자 인덱스를 인용해 중산산의 재산이 778 억 달러약 84 조 6464 억원를 기록했다고 지난달 31 일현지시간 보도했다. 이는 세계에서는 11 위, 아시아에서는 1 위에 해당하는 재산 규모다. 중산산의 재산은 2020 년 한 해에만 709 억 달러약 77 조 1392 억원가 늘었다. 블룸버그는 역사상 재산이 가장 빠르게 늘어난 경우 중 하나라며 올해까지 그가 중국 외에는 거의 알려지지 않았다는 점을 고려하면 주목할 만하다고 밝혔다. 중산산의 성공 요인으로는 우선 지난해 농푸산취안과 백신 제조업체 완타이바이오의 상장이 꼽힌다. 상장 이후 농푸산취안의 주가는 155 올랐고, 완타이바이오는 무려 2000 이상 급등했다. 또 중산산은 별명이 외로운 늑대인 대로 중국 정치에 관여하지 않고, 그의 사업은 중국 내 다른 부호들과도 얽혀있지 않는다는 평가도 받고 있다. 반면 한때 아시아와 중국의 최고 부자였던 마윈 알리바바 창업자는 최근 중국 정치권을 비판했다가 재산이 크게 줄었다. 당국의 전방위 규제 압박을 받은 마윈은 앤트그룹 상장이 무산될 위기에 처하는 등 #idiom#을 겪으면서 617 억 달러약 67 조 1296 억원였던 재산이 512 억 달러약 55 조 7056 억원까지 줄어들었다. 한편 중산산에게 아시아 부호 1 위를 내준 인도 재벌 암바니 회장의 재산은 2020 년 한 해 동안 183 억 달러약 19 조 9104 억원가 늘어난 769 억 달러약 83 억 6672 억원로 집계됐다. <br><br> Became Asia's No. 1 by selling water, surpassing Jack Ma. Bloomberg reported on the 31st (local time) that Zhong Shanshan's wealth has reached $77.8 billion (about 84.6464 trillion KRW), citing its Billionaires Index. This ranks 11th in the world and 1st in Asia. Zhong's wealth increased by $70.9 billion (about 77.1392 trillion KRW) in 2020 alone. Bloomberg highlighted this as one of the fastest wealth increases in history, noting that Zhong was virtually unknown outside of China until this year. The key to his success is attributed to the IPOs of Nongfu Spring and vaccine manufacturer Wantai Bio in the previous year. After the IPO, Nongfu Spring's stock price rose by 155%, while Wantai Bio's stock price surged by over 2000%. Furthermore, Zhong is nicknamed the "Lone Wolf" and is noted for not being involved in Chinese politics, with his business having no ties to other Chinese billionaires. On the other hand, Jack Ma, the founder of Alibaba and once the richest man in Asia and China, saw his wealth significantly shrink after criticizing Chinese politics. Facing widespread regulatory pressure from authorities and the risk of the Ant Group's IPO being canceled, he experienced #idiom#, with his wealth dropping from $61.7 billion (about 67.1296 trillion KRW) to $51.2 billion (about 55.7056 trillion KRW). Meanwhile, Indian tycoon Mukesh Ambani, who lost his spot as Asia's richest to Zhong, saw his wealth increase by $18.3 billion (about 19.9104 trillion KRW) to a total of $76.9 billion (about 83.6672 trillion KRW) in 2020. |
|---|---|
| **Model Choice (From GPT-3.5 Turbo)** | 삼고초려 <br> (Patiently striving to attract talent) |
| **Golden Answer** | 우여곡절 (ups and downs) |
| **Model Response** | The idiom '삼고초려' means to carefully consider and think deeply before making a decision. In the context of the text, it is the best choice for replacing #idiom# because it highlights the importance of careful consideration and decision-making in managing wealth and business ventures, as seen in the success of Zhong Shanshan who carefully chose his investments and avoided political entanglements, in contrast to Jack Ma who faced setbacks due to his outspoken criticism of the Chinese government. This idiom emphasizes the significance of thoughtful planning and strategic choices in achieving and maintaining financial success. |

Table 8 An example of lack of cultural background knowledge in model responses

**A logical error** is defined when the response provided by a model is inconsistent or contradictory in terms of logic. This type of error reflects a breakdown in reasoning. For example, in the sample shown in Table 10, this model identified the correct option, but made a contradictory choice when deciding on the answer, despite recognizing the correct option.

| | |
|---|---|
| News Text | 고정주는 구한말의 애국계몽운동가이다. 그의 자는 보현, 호는 춘강이다. 담양 창평에서 1863 년에 참봉을 지낸 아버지 고제두와 어머니 전주 이씨 사이에서 태어났다. 5 세 때 큰아버지인 선공감 감역 고제승의 양자로 들어가 양부로부터 학문을 배우고 13 세부터 지금의 상월정에서 열심히 공부하였다. 19 세 때 한장석을 찾아가 학문을 배웠으며 21 세에는 성대영을 찾아가 가르침을 받았는데 이 때 그의 학문이 높은 수준에 이르러 성대영이 높이 칭찬했다고 한다. 그의 스승인 한장석은 정치적 실천을 중시했던 인물로 경세학을 강조하였다. 이른바 실천하지 않는 지성은 시대의 방관자이며 무책임한 선비임을 깨달았기에 고정주는 관직에 있었던 시절이나 귀향하여 신교육운동에 몰입했던 시절에도 스승의 가르침을 깊이 새기며 실천하는 지성인의 모범을 보여주었다. 1885 년에 진사에 합격하고 1891 년에 문과에 합격하였다. 이 때 동생도 같이 진사에 합격하여 형제간에 금의환향하였다. 1893 년 승문원 부정자로 관직을 시작하였는데 고종도 크게 관심을 보이며 제봉 고경명의 몇 대손인가를 묻고 선물을 하사했다고 한다. 1898 년 종묘축관에 선출되었고, 1899년에는 홍문관 시독을 역임하고 품계가 정 6 품인 승훈랑에 올랐다. 그러나 이 때는 외세의 침탈과 내부의 혼란으로 나라가 풍전등화의 위기에 처한 시기였다. 이에 고정주는 #idiom#을 강조하는 상소를 올렸다. 이 상소에서 그는 상벌을 명백히 하여 부패한 관리를 추방하고 재원을 모아 군사력을 증대하여 국가의 근본을 굳건하게 함으로써 외세의 침입을 물리칠 것을 강력히 주장하였지만 당시 정치상황은 그의 주장을 받아들이지 못했다. (English Summary: Go Jeong-ju was a patriotic enlightenment activist during the late Joseon Dynasty. His courtesy name was Bohyeon, and his pen name was Chungang. Born in 1863 in Changpyeong, Damyang, to Go Je-du and Lady Jeon from the Jeonju Yi clan, Go was adopted by his uncle Go Je-seung at age five, from whom he began his academic studies. He later continued his education under prominent scholars like Han Jang-seok and Seong Dae-young, who praised Go's intellectual abilities. Han Jang-seok, Go's mentor, emphasized practical application of knowledge, which deeply influenced Go. Go passed the civil service examination in 1885 and began his career in 1893, with King Gojong recognizing his potential. Despite his rise in rank, Go witnessed Korea's increasing vulnerability to foreign powers and internal instability. He submitted a memorial to the king, emphasizing #idiom#, calling for the expulsion of corrupt officials, the strengthening of military forces, and the consolidation of national resources to resist foreign invasion. However, his proposals were not accepted in the political climate of the time.) |
| Model Choice (From Tiangong) | 内수외양 |
| Golden Answer | 내수외양(Strengthen internal governance and build external power.) |
| Model Response | This article discusses GO JEONGJU's patriotic ideology. He advocated for consolidating the nation's foundation to address the crises of foreign invasions and internal turmoil. His ideas and practices had a profound impact on society at the time. Among the provided options, the idiom that best fits into the text at the position marked as #idiom# is "内수외양." This idiom means to strengthen governance internally and enhance power externally, aligning with the theme emphasized by GO JEONGJU. Therefore, I choose "内수외양" as the most suitable idiom. [내수외양] |

Table 9 An example of grammatical or lexical error in model responses

**Insufficient context interpretation** refers to the failure of a model to adequately understand and utilize the broader context or cultural nuances of a text. This error manifests particularly in settings where emotional resonance and cultural specificity are crucial, such as in poetry or other richly emotional texts. When a model exhibits insufficient context interpretation, it might produce content that, while technically correct, does not align emotionally or culturally with the intended message as shown in Table 11.

In the expression "하늘을 우러러", the poet is metaphorically expressing a desire to pursue an ideal realm. However, the speaker's feet remain planted on the ground, and there is sadness due to the inability to fully extend them. The speaker is experiencing a world where reality and ideals coexist, feeling a cognitive paradox. The model's elected answer "나 두 야 간다", while grammatically correct and forming a complete sentence, completely overlooks the emotional and cultural background of the poem. This error demonstrates the model's limitations in handling texts that require an understanding of extensive cultural and emotional contexts.

| | |
|---|---|
| News Text | 추억의 보물을 어제 찾았다. 흰 머리칼을 날리며 유년기를 재현했다. 평생 머릿속에 남아있던 보물찾기를 실시한 야유회는 잊을 수 없다. 놀이하면 보물찾기와 수건돌리기가 떠오르니 아직도 철이 안든 모양이다. 이번 강원수필 하계 야유회에서도 그런 나의 아우성을 받아들여 프로그램에 선뜻 넣어 고마웠다. 춘천 인근이었다. 싱그러운 신록에 푹 파묻혀 #idiom#하며 풀벌레소리에 졸고 있는 체험농원을 찾았다. 임시회의를 열어 두 개의 안건을 통과시켰지만, 나에겐 소년처럼 보물찾기가 마치 연민의 정과도 같았다. 회의가 끝나고 점심도 마치고, 드디어 농원이 거느린 정원 둘레에 15 개의 보물이 숨어있다고 과거시험 시제처럼 일러준다. 회원 대부분이 고희의 포구에 정박하고 있었지만, 누구 하나 마다하지 않고 서둘러 숨겨놓은 곳으로 달려 나가기에 바빴다… (English Translation: I found the treasure of my memories yesterday, recreating my childhood with white hair blowing in the wind. I'll never forget the picnic where we played treasure hunt, which had stayed in my mind all my life. When I think of games, treasure hunts and 'hide and seek' come to mind, making me realize I haven't quite grown up yet. I'm grateful that at the Gangwon Essay summer picnic, they willingly added the treasure hunt to the program at my request. It was near Chuncheon. Immersed in lush greenery, #idiom#, we visited an experiential farm where we dozed off to the sounds of insects. Although we held a temporary meeting and passed two resolutions, to me, the treasure hunt felt like a nostalgic yearning, something akin to a boy's adventure. After the meeting and lunch, we were finally told, as if it were a state exam, that 15 treasures were hidden around the farm's garden. Although most members had docked at the harbor of their 70s, no one hesitated, and everyone eagerly rushed to search for the hidden treasures...) |
| Model Choice (From Clova X) | 천금매소(to spend a fortune to buy a smile.) |
| Golden Answer | 고침안면 (to sleep peacefully) |
| Model Response | The choice of the is idiom "천금매소" because, in these precious moments of reminiscence, we relive the pure joy of childhood, akin to buying a moment's laughter with a great sum of money. Although the correct idiom should have been "고침안면", considering the relationship we have with the treasures in the game, it seems like we are buying back the joy of our childhood with money, hence the selection of "천금매소" is more symbolic. |

Table 10 An example of logical error in model responses

| | |
|---|---|
| **Poem** | #sentence#<br>울기는 하여도<br>하늘이 그리워 울음이 아니다<br>두 발을 못 뻗는 이 땅이 애달파<br>하늘을 흘기니<br>울음이 터진다.<br>해야 웃지 마라.<br>달도 뜨지 마라.<br>#sentence#<br>Even if I cry,<br>It is not because I miss the sky.<br>It is the sorrow of this land,<br>Where my feet cannot stretch.<br>Looking up at the sky,<br>Tears break forth.<br>Oh sun, do not laugh.<br>Moon, do not rise either. |
| **Model Choice (From Clova X)** | 나 두 야 간다 (I too shall leave) |
| **Golden Answer** | 하늘을 우러러 (Looking up at the sky) |
| **Model Response** | The sentence '나 두 야 간다' is the best choice because it is a complete and grammatically correct sentence in Korean. It translates to "I am also going" in English. The sentence includes the subject '나' (I), the adverb '도' (also), and the verb '가다' (to go), making it a simple and clear statement. Additionally, the sentence follows the standard word order in Korean, which is subject-object-verb. Overall, '나 두 야 간다' effectively conveys the message of the speaker joining someone else in going somewhere. |

Table 11 An example of insufficient context interpretation in model responses